\documentclass[11pt]{article}

\usepackage{acl}  

\usepackage{times}
\usepackage{latexsym}

\usepackage[ruled, lined, linesnumbered, commentsnumbered]{algorithm2e}
\usepackage{amsmath,amssymb}

\usepackage[T1]{fontenc}

\usepackage[utf8]{inputenc}

\usepackage{microtype}

\usepackage{inconsolata}

\usepackage{graphicx}

\usepackage{amsmath}
\usepackage{quoting}
\usepackage{hyperref}
\usepackage{todonotes}
\usepackage{mdframed}
\usepackage{booktabs} 
\usepackage{tabularx} 
\usepackage{multirow} 
\usepackage{listings}
\usepackage[T1]{fontenc}
\usepackage{siunitx}

\usepackage[no-math]{xeCJK} 
\setCJKmonofont{NotoSansMonoCJKjp-Regular.otf}
\newfontfamily\unicodefont{FreeSerif.otf}

\newcommand{\llmoutput}[1]{%
    \small\ttfamily\unicodefont
    \begin{minipage}[t]{\linewidth}
    \vspace{2pt}
    \raggedright #1
    \vspace{4pt}
    \end{minipage}
}

\usepackage{amsmath,amsfonts,bm}

\def\eqref#1{equation~\ref{#1}}

\def\1{\bm{1}}

\def\vzero{{\bm{0}}}

\def\vg{{\bm{g}}}

\def\vv{{\bm{v}}}

\DeclareMathAlphabet{\mathsfit}{\encodingdefault}{\sfdefault}{m}{sl}
\SetMathAlphabet{\mathsfit}{bold}{\encodingdefault}{\sfdefault}{bx}{n}

\def\gN{{\mathcal{N}}}

\newcommand{\Cin}{\ensuremath{\mathcal{C}_{\text{in}}}}
\newcommand{\Cout}{\ensuremath{\mathcal{C}_{\text{out}}}}

\newcommand{\ftparams}{\ensuremath{\theta_\textrm{ft}}}
\newcommand{\ptparams}{\ensuremath{\theta_\textrm{pt}}}

\title{Advancing the State-of-the-Art in Empirical Privacy Auditing}

\author{
  Nicole Mitchell\thanks{NM was primarily responsible for experiments, analysis, and presentation of results. GA was primarily responsible for conceiving of the synthetic canary design and the LLM-based data audit, and paper writing. Both contributed to implementation.}\\
  Google Research \\
  San Francisco, CA \\
  \texttt{nicolemitchell@google.com} \\\And
  Galen Andrew\footnotemark[1] \\
  Google Research \\
  Seattle, WA \\
  \texttt{galenandrew@google.com} \\
  \AND
  Arun Ganesh \\
  Google Research \\
  Seattle, WA \\
  \texttt{arunganesh@google.com} \\\And
  Brendan McMahan \\
  Google Research \\
  Seattle, WA \\
  \texttt{mcmahan@google.com} \\\And
  Peter Kairouz \\
  Google Research \\
  Seattle, WA \\
  \texttt{kairouz@google.com}\\\\
}

\begin{document}
\maketitle

\begin{abstract}

Parameter-efficient fine-tuning of large language models (LLMs) can exhibit problematic memorization of individual training examples.  Empirical privacy auditing (EPA) quantifies this risk by measuring realistic data leakage on membership inference (MI) or reconstruction attacks. A key challenge in EPA is designing ``canary'' examples that are mixed with the privacy-sensitive training data.  We propose generating synthetic canaries via high-temperature sampling ($T \geq 0.8$) from LLMs, using prompts tailored to the privacy-sensitive training data. These canaries act as high-influence outliers, ensuring high identifiability and hence strong audits. Further, since the canaries are themselves non-private, they are inspectable and can be inserted with repetition without jeopardizing the privacy of the real data. An important use of models fine-tuned on privacy-sensitive data is the generation of synthetic data. This also comes with privacy risk. We introduce a powerful synthetic data audit based on fine-tuning an auxiliary model on the synthetic data. Auditing the auxiliary model for the original canaries then provides a strong estimate of the privacy leakage through the synthetic data. Finally, leveraging our strong auditing methodologies, we perform a systematic investigation into the interacting effects of model capacity and canary entropy on memorization.
\end{abstract}

\section{Introduction} \label{sec:intro}

Large Language Models (LLMs) are increasingly fine-tuned on sensitive, domain-specific data, raising significant concerns regarding the memorization and subsequent leakage of training records \cite{kandpal2024user}.  To quantify these practical deployment risks, Empirical Privacy Auditing (EPA) has become an essential evaluation tool. EPA is often operationalized by executing Membership Inference (MI) or data reconstruction attacks against a target model \citep{shokri2017membership, jagielski2020auditing, carlini2021extracting, nasr2021adversary, carlini2022membership}. To the extent that the model withstands strong simulated attacks, we have increased confidence that the model can be deployed without risking exposure of sensitive information inherent in the training data.

A standard EPA procedure involves injecting traceable ``canaries'' into the training set and subsequently attacking the trained model (or the synthetic data it generates) to determine if the canary's presence can be detected or its contents extracted~\citep{carlini2019secret}. The success of the attack depends heavily on the specific nature and design of these canaries, among other factors~\citep{nasr2023tight}. 

Ideal canaries must balance a tradeoff between being in- vs.\ out-of-distribution with respect to the private fine-tuning data. On the one hand, out-of-distribution canaries could act as high-influence outliers to maximize the privacy leakage signal, giving empirical estimates that accurately reflect the risk of a realistic attack against worst-case (most easily identifiable) data~\citep{maddock2022canife, koskela2025auditing}. However, examples which are too far out-of-distribution may not be learnable with the limited capability of parameter-efficient fine-tuning (LoRA)~\citep{hu2022lora}, or may not be represented in synthetic data generated by a model \citep{meeus2025canarysecho}. Another consideration is that auditing with canaries that are closer to the private data distribution may be more effective at providing reassurance to users that data ``like theirs'' is not vulnerable.

In this paper, we propose a novel canary generation method that resolves the tension between identifiability and naturalness. We introduce the use of synthetic canaries generated by prompting a pretrained, instruction-tuned LLM at a high temperature ($T \geq 0.8$)~\citep{ackley1985learning}. Scaling the sampling temperature flattens the output distribution, producing text that is statistically unlikely under the base distribution but usually retaining essential syntactic and semantic structures. Our canaries are sufficiently unusual to trigger a measurable response during gradient updates, rendering them highly susceptible to MI and reconstruction attacks, while remaining sufficiently similar to real user data and structurally coherent enough to avoid corrupting the model's primary objective.

A crucial advantage of our approach is that it makes no reference to the private data beyond knowledge of the basic format in order to design an effective prompt. Thus, it is unnecessary to account for the privacy cost of accessing the data for canary generation. Group privacy properties (security of secrets shared by small groups) can be ascertained by inserting canaries with repetition, which would otherwise increase vulnerability of the repeated user data. In addition, it can be convenient to be able to freely inspect the canaries during development without compromising user privacy.

We also introduce a novel methodology for auditing synthetic data. Existing data audits generally assume that leakage appears as verbatim regurgitation or direct semantic similarity between canaries and individual synthetic examples generated by the model~\citep{meeus2025canarysecho}. In practice, a canary can exert a more subtle, diffuse influence, slightly shifting the distribution of tokens or concepts across the entire synthetic dataset~\citep{cloud2026language}. To capture this sparse signal, our audit fine-tunes an auxiliary ``attack'' LLM on the generated synthetic data. Because the attack model aggregates these diffuse signals into its weights, a canary observed during training of the primary model will exhibit a higher likelihood (according to the attack model) than an unobserved canary from the same distribution. 

This proxy-based approach effectively bridges data and model auditing, allowing us to run standard model-level MI attacks on the attack model to detect subtle data leakage. 
Unlike prior approaches, we directly audit the concrete set of synthetic data to be released as opposed to requiring generating multiple shadow sets in a Robust-MIA style attack~\cite{zarifzadeh2024lowcost}. 
Furthermore, because we can use exactly the same auditing approach for the original fine-tuned model and the auxiliary model, we gain a fair assessment of the relative leakage of releasing a fine-tuned model, versus only releasing synthetic data sampled from it.

The primary contributions of our work are as follows:
\begin{enumerate}
    \item We introduce high-temperature sampling as a scalable, automated method to generate highly effective canaries for LLM auditing. These synthetic canaries are somewhat in-distribution for the pre-trained model, and hence easy for the pre-trained model to memorize even via LoRA. Thus they yield significantly stronger MI and reconstruction attack signals for LoRA compared to baselines from the literature.
    \item We design a new data auditing methodology to estimate leakage in synthetic data generated by a fine-tuned model. We demonstrate that this attack is significantly more sensitive across canary types than baseline attacks from the literature.
    \item Using our canaries and powerful model and data audits, we perform a systematic study of the effect of model capacity (as measured by LoRA rank) and canary entropy on memorization metrics. 
    We find that the canary entropy that maximizes memorization increases with LoRA rank,
    and is lower for reconstruction attacks than MI attacks.
\end{enumerate}

\section{Preliminaries and Related Work}
\subsection{Differentially Private Training}

Differential privacy (DP) provides a rigorous framework for formal privacy protections. While DP guarantees can be expressed in several ways, with approximate $(\varepsilon, \delta)$-DP being particularly well-known in the ML community, we follow \citet{gomez2025gaussian} in focusing on Gaussian Differential Privacy ($\mu$-GDP)~\citep{dong2022gaussian}, as it often more accurately expresses the privacy profile of modern DP machine learning algorithms. Two datasets $D$ and $D'$ are considered adjacent if they differ by the addition or removal of a single example $x$, i.e., $D' = D \cup \{x\}$. A randomized algorithm $\mathcal{M}$ satisfies $\mu$-GDP if for any adjacent $D, D'$ distinguishing between the output distributions $\mathcal{M}(D)$ and $\mathcal{M}(D')$ is at least as statistically difficult as distinguishing between two normal distributions $\mathcal{N}(0, 1)$ and $\mathcal{N}(\mu, 1)$. This provides a single, highly interpretable privacy parameter $\mu$, where a smaller $\mu$ corresponds to stronger privacy protection. DP-SGD~\citep{song2013stochastic,abadi16deep} is the canonical algorithm for training machine learning models with DP. A full description of DP-SGD is given in App.~\ref{apx:dp}. Our primary results examine the memorization properties of non-DP models, but we still use an empirically-estimated $\mu$ to quantify memorization as we describe next.

\subsection{Empirical Privacy Auditing} \label{sec:epa}

Empirical privacy auditing refers to a suite of methods to estimate the practical privacy risk of training on sensitive data (c.f. \citet[Sec 7.5]{ponomareva2025dpfydata}). We instantiate a hypothetical adversary with at least as much power and information as a realistic adversary might be presumed to have. In this work we consider several attacks that vary the adversary's information and goal. In a \emph{model audit}, the adversary has access to the fine-tuned model, whereas in a weaker \emph{data audit}, they only observe a finite set of synthetic examples generated from the model. We also consider two potential adversary goals: \emph{reconstruction}, in which the adversary attempts to infer some unknown text (the suffix) given a known prefix, and \emph{membership inference}, in which the adversary has the relatively easier task of guessing whether a known example was part of training or not. Both attacks require a set of canaries $\Cin$ that are inserted into the training data (potentially with repetition), and another set $\Cout$ drawn from the same distribution that functions as a control. Denote by $\ptparams$ the pre-trained model parameters (used to initialize fine-tuning) and by $\ftparams$ the model parameters after fine-tuning with canaries inserted.

\paragraph{Reconstruction attack.} A straightforward way to attack an LLM is to assume the adversary knows the prefix of some user's data, and then prompt the model with that prefix with greedy decoding to attempt to learn the suffix. In our setup, we take the first 60 tokens of a canary, and ask the model to complete the 10-token suffix given the first 50 tokens.  The adversary ``wins'' if the Levenshtein edit distance between the completion and the true suffix is less than $d$ (where we consider $d$ from 0 to 5). Because for some canary types the adversary could guess a few tokens purely based on the statistics of language, we measure the \emph{lift}, $L_d$, or how many \emph{more} seen canary suffixes are completed at a given edit distance compared to unseen canary suffixes:
\[ \delta(c) = \textrm{edit} (\textrm{greedy}(c\texttt{[:-10]}\ |\ \ftparams),\, c\texttt{[-10:]}) \]
\[L_d = \left| { c \in \mathcal{C}_\text{in} : \delta(c) \leq d } \right|- \left| { c \in \mathcal{C}_\text{out} : \delta(c) \leq d } \right|\]
\paragraph{Membership Inference attack.} Here the adversary's goal is to guess whether a known-verbatim canary was part of training or not. We quantify the success of a membership inference attack by estimating its empirical $\mu$-GDP parameter from the likelihood ratio scores of held-in vs.\ held-out canaries: \[s(c) = \log \Pr(c\ |\ \ftparams) - \log \Pr(c\ |\ \ptparams). \] The positive term is precisely what generative fine-tuning attempts to maximize, while the negative term adjusts for the intrinsic likelihood of the canary text. We report the empirical $\mu$-GDP~\citep{nasr2023tight, mahloujifar25empirical} of the attack. Recent work shows that for realistic attacks $\mu$-GDP estimates are relatively stable across classification thresholds, except for the extreme tails where noise dominates~\citep{koskela2025auditing}. Defining the number of true positives and true negatives for a given threshold as
\begin{align*} n_\mathrm{tp}(\tau) &= \left| \{ c \in \Cin : s(c) > \tau \} \right| \\ n_\mathrm{fp}(\tau) &= \left| \{ c \in \Cout : s(c) > \tau \} \right|,
\end{align*}
we consider the set of thresholds with at least 30 examples classified as positive or negative:
\[ \mathcal{T} = \{ \tau: 30 \leq n_\mathrm{fp}(\tau) + n_\mathrm{tp}(\tau) \leq 2n - 30 \}, \]
where $n = \| \Cin \| = \| \Cout \|$ is the number of unique canaries per set. We estimate the true positive and false positive rates using a Jeffreys prior so our $\mu$ estimate reaches a large but bounded value even when the canaries are perfectly identifiable:\footnote{In a real privacy auditing scenario, all models in this low privacy regime should be considered unsafe, so this smoothing would not matter -- we could just report an infinite estimate when $n_\mathrm{tp} = 0$, for example. But, it is useful in this work to be able to distinguish between settings with high memorization.}
\[ r_{\mathrm{fp}}(\tau) = \frac{n_\mathrm{fp}(\tau) + 0.5}{n + 1}, r_{\mathrm{tp}}(\tau) = \frac{n_\mathrm{tp}(\tau) + 0.5}{n + 1}. \]
Finally we estimate $\mu$ according to
\[ \mu = \max_{\tau \in T} \left| \Phi^{-1}(r_\mathrm{tp}(\tau)) - \Phi^{-1}(r_\mathrm{fp}(\tau)) \right|, \]
where $\Phi^{-1}$ is the inverse CDF of the standard normal distribution. We use $n = 3000$ unique canaries per set, so the estimated $\mu$ of a perfect classifier is 7.17. We use bias corrected and accelerated (BCa) bootstrapping over the sets of scores to estimate a 95\% confidence interval on $\mu$~\citep{efron1987better}. In addition, we report the common metric of TPR-at-fixed-low-FPR: $r_{\mathrm{tp}}(\tau^*)$ where $r_\mathrm{fp}(\tau^*)=0.01$ or $0.1$.

\subsection{Synthetic Data Auditing} \label{sec:data-audit}
Private synthetic text generation---fine-tuning a model with DP over private data and prompting it to sample from the learned distribution---is a useful way to produce high-fidelity linguistic data tuned to a target distribution without exposing the sensitive information~\citep{ponomareva2025dpfydata}.

The goal of synthetic data auditing is to trace the influence of individual training examples through to a synthetic dataset $\mathcal{D} = \{d_i\}$ generated by the fine-tuned model $\ftparams$. Having access only to $\mathcal{D}$ makes the attacker's task much harder, but as we will show, strong membership inference and even significant reconstruction is still possible. \citet{meeus2025canarysecho} introduce two forms of data-based MI attacks. Like the model attack described in Sec. \ref{sec:epa}, both derive a membership signal $s(c)$, but $s$ now depends only on $\mathcal{D}$ and not directly on $\ftparams$. In the similarity-based attack, the attacker computes the similarity of the canary to each synthetic example $\sigma_i(c) = \text{SIM}(c, d_i)$, using for example Jaccard similarity for strings or cosine similarity between embeddings. Then the score of a canary is the mean similarity of the top $k$ most-similar examples: $s_\text{sim}(c) = \frac{1}{k} \sum_{j=1}^k \sigma_{i(j)}$.\footnote{\citet{meeus2025canarysecho} find the similarity attacks to be weaker than the bigram attack, so we do not compare to them in this study.} In the bigram attack, the attacker trains a simple bigram language model with Laplace smoothing over the synthetic data and uses it to estimate the likelihood of the canary. Thus
\[ s_\text{bi}(c) = \sum_{i=1}^L \log \frac{\# (c\texttt{[i-1]}, c\texttt{[i]}) + 1}{\# c\texttt{[i-1]} + V},  \]
where $\# t$ represents the number of occurrences of a unigram or bigram in $\mathcal{D}$, and $V$ is the size of the vocabulary.\footnote{\citet{meeus2025canarysecho} generate multiple sets of data to get a set of ``in'' and ``out'' scores for each canary in an RMIA-style attack~\citep{zarifzadeh2024lowcost}. Our method (described in Sec.~\ref{sec:model-based-data-audit}) requires only a single synthetic dataset (the one that is to be certified private for release), so we also only compute our bigram language model over that single dataset for this attack. Note that we could also apply RMIA to our attack, which should also make it stronger.}

\subsection{Related Work}


Reconstruction and membership inference attacks such as the ones we use are now the de-facto methods for privacy auditing \citep{shokri2017membership, hayes2017logan, carlini2019secret,carlini2021extracting}. While $(\varepsilon, \delta)$-DP has often been used as the measure of leakage in privacy auditing, $\mu$-GDP \citep{dong2022gaussian} has recently gained popularity as a leakage measure \citep{mahloujifar25empirical, koskela2025auditing, nasr2023tight} because it lends to tighter audits and better aligns with the hypothesis testing (TPR vs. FPR) notion of differential privacy \citep{wasserman08statistical, kairouz2015composition} used in auditing works, even those not focused on differential privacy.

A central design choice in empirical auditing is the construction of the canaries themselves. In the context of auditing models, \citet{carlini2019secret} introduced canaries as inserted sequences for measuring unintended memorization. Since then, canary design has become an active research area: Canife uses gradient-based optimization to produce highly detectable canaries \citep{maddock2022canife}; \citet{boglioni2026optimizing} optimize canaries using meta-gradient descent for an image classification task;  other work has used fixed templated structures \citep{nasr2023tight}; and \citet{panda2025privacy} proposed least-common-bigram canaries as an out-of-distribution construction for privacy auditing. The most closely related work is \citet{meeus2025canarysecho}, which studies privacy auditing for synthetic data released by a fine-tuned model. Their canaries combine prefixes sampled from the private data with generated suffixes, and their data audits use whole-example similarity-based and bigram-based attacks.

We improve on prior state-of-the-art in several ways. First, as our canaries are generated by prompting the pretrained model, they tend to be more easily memorizable when fine-tuning that model (as we demonstrate). They also do not require access to the private training data. At the same time, an attack using our synthetic canaries is more plausible compared to canaries constructed from unseen bigrams, or out-of-vocabulary tokens that could not occur in real user data. Second, our model-based synthetic data audit lets diffuse leakage signals accumulate in the attack model rather than requiring direct verbatim overlap at the example or even bigram level, significantly increasing its generality and power. Similar to \citet{marek2026benchmarking}, we benchmark our canaries and data auditing method in the context of low rank adaptation. Our findings on the memorization properties of LoRA echo their results: LoRA offers more protection for OOD data~\citep{marek2026benchmarking}, see App.~\ref{apx:rank_ablation}.

\begin{algorithm*}[h]
\caption{Fine-tune model with Canaries}
\label{alg:finetune}
\KwIn{Pretrained parameters $\ptparams$; Fine-tuning dataset $\mathcal{D}$;
      canaries $\Cin$; repetitions $r$;
      steps $t$; learning rate $\eta$; batch size $B$;
      DP flag; noise multiplier $\sigma$; clip norm $S$;
      Poisson expected batch size $q \cdot |\mathcal{D}_\textrm{Enron}|$}
\KwOut{Fine-tuned parameters $\ftparams$}

\tcp{Inject canaries into training data}
$\mathcal{D}' \leftarrow \mathcal{D} \setminus \text{last } r|\Cin| \text{ examples}$\;
$\mathcal{D}' \leftarrow \textsc{Shuffle}\!\left(\mathcal{D}' \cup \Cin^{\times r}\right)$\;

\tcp{Initialize trainable parameters (LoRA adapter or full weights)}
$\ftparams \leftarrow \textsc{InitParams}(\ptparams,\; \text{lora\_rank})$\;

\For{step $t = 1$ \KwTo $t$}{
    \eIf{DP}{
        $\mathcal{B} \sim \textsc{PoissonSample}(\mathcal{D}',\, q)$
            \tcp*{expected size $q|\mathcal{D}'|$}
        $g \leftarrow \frac{1}{|\mathcal{B}|}\!\sum_{x \in \mathcal{B}}
            \textsc{Clip}\!\left(\nabla_{\ftparams}\,
            \mathcal{L}(x;\ftparams),\; S\right)
            + \mathcal{N}(0,\,\sigma^2 S^2 \mathbf{I})$\;
    }{
        $\mathcal{B} \sim \textsc{MinibatchSample}(\mathcal{D}',\, B)$\;
        $g \leftarrow \frac{1}{|\mathcal{B}|}\!\sum_{x \in \mathcal{B}}
            \nabla_{\ftparams}\, \mathcal{L}(x;\ftparams)$\;
    }
    $\ftparams \leftarrow \textsc{AdamW}(\ftparams,\, g,\, \eta)$\;
}
\Return $\ftparams$\;
\end{algorithm*}

\begin{algorithm*}[h]
\caption{Audit a Fine-tuned Model}
\label{alg:audit}
\KwIn{Pretrained parameters $\ptparams$;
      fine-tuned parameters $\ftparams$;
      canaries $\Cin$ (seen), $\Cout$ (unseen);
      canary token length $\ell = 60$}
\KwOut{Estimated $\hat\mu$; reconstruction lift $\{L_d\}_{d=0}^{5}$}

\tcp{Membership inference: likelihood-ratio score (§2, Eq.~s(c))}
\ForEach{$c \in \Cin \cup \Cout$}{
    $s(c) \leftarrow \log \Pr\!\left(c[{:}\ell] \mid \ftparams\right)
                   - \log \Pr\!\left(c[{:}\ell] \mid \ptparams\right)$\;
}
$\hat\mu \leftarrow \textsc{EstimateGDP}\!\left(
    \{s(c)\}_{c \in \Cin},\;
    \{s(c)\}_{c \in \Cout}\right)$\;

\tcp{Reconstruction attack: greedy suffix completion (§2, Eq.~$\delta(c)$)}
\ForEach{$c \in \Cin \cup \Cout$}{
    $\delta(c) \leftarrow
        \textrm{edit}\!\left(
            \textrm{greedy}(c\texttt{[:-10]} \mid \ftparams),\;
            c\texttt{[-10:]}\right)$\;
}
\For{$d = 0$ \KwTo $5$}{
    $L_d \leftarrow \left|\{c \in \Cin : \delta(c) \leq d\}\right|
                  - \left|\{c \in \Cout : \delta(c) \leq d\}\right|$\;
}

\Return $\hat\mu$, $\{L_d\}_{d=0}^{5}$\;
\end{algorithm*}

\begin{algorithm*}[h]
\caption{Generate Synthetic Data from a Fine-tuned Model}
\label{alg:generate}
\KwIn{Fine-tuned parameters $\ftparams$; prompt $p$;
      number of examples $N$;
      temperature $T$; top-$p$ nucleus $p_\text{nuc}$}
\KwOut{Synthetic dataset $\mathcal{D} = \{d_i\}_{i=1}^{N}$}

$\mathcal{D} \leftarrow \varnothing$\;
\While{$|\mathcal{D}| < N$}{
    $\mathcal{B} \leftarrow \textsc{Sample}\!\left(\ftparams,\;
        \text{input} = p,\; T,\; p_\text{nuc}\right)$
        \tcp*{autoregressive generation}
    $\mathcal{D} \leftarrow \mathcal{D} \cup \mathcal{B}$\;
}
\Return $\mathcal{D}$\;
\end{algorithm*}

\begin{algorithm*}[h]
\caption{Full Privacy Auditing Pipeline}
\label{alg:pipeline}
\KwIn{Pretrained parameters $\ptparams$;
      Fine-tuning dataset $\mathcal{D}$;
      canary sets $\Cin,\, \Cout$;
      hyperparameters (Algorithms~\ref{alg:finetune}--\ref{alg:generate})}
\KwOut{Model audit and synthetic data audit results}

\tcp{\textbf{Stage 1}: Fine-tune on original data with canaries}
$\ftparams \leftarrow
    \textsc{FineTune}(\ptparams,\; \mathcal{D},\; \Cin)$
    \hfill (Algorithm~\ref{alg:finetune})\;

\tcp{\textbf{Stage 2}: Audit the fine-tuned model}
$\hat\mu_\text{model},\; \{L_d\}_\text{model} \leftarrow
    \textsc{Audit}(\ptparams,\; \ftparams,\; \Cin,\; \Cout)$
    \hfill (Algorithm~\ref{alg:audit})\;

\tcp{\textbf{Stage 3}: Generate synthetic data from $\ftparams$}
$\mathcal{D}_\textrm{synth} \leftarrow
    \textsc{GenerateSynthetic}(\ftparams)$
    \hfill (Algorithm~\ref{alg:generate})\;

\tcp{\textbf{Stage 4}: Fine-tune attack model on synthetic data with canaries}
$\theta_\textrm{atk} \leftarrow
    \textsc{FineTune}(\ptparams,\; \mathcal{D}_\textrm{synth},\; \Cin)$
    \hfill (Algorithm~\ref{alg:finetune})\;

\tcp{\textbf{Stage 5}: Audit the attack model}
$\hat\mu_\text{data},\; \{L_d\}_\text{data} \leftarrow
    \textsc{Audit}(\ptparams,\; \theta_\textrm{atk},\; \Cin,\; \Cout)$
    \hfill (Algorithm~\ref{alg:audit})\;

\Return
    $\bigl(\hat\mu_\text{model},\; \{L_d\}_\text{model}\bigr)$,\quad
    $\bigl(\hat\mu_\text{data},\; \{L_d\}_\text{data}\bigr)$\;
\end{algorithm*}

\section{Synthetic Canary Design}
\label{sec:canary_design}

We generate synthetic canary examples by prompting the instruction-tuned version of the same pre-trained model that will be used to initialize our audited fine-tuning run\footnote{We experiment with using a different model to generate canaries in App.~\ref{apx:model_ablation}.}. In our experiments we use the Gemma3 12B-parameter model, and fine-tune it on the Enron email corpus. We propose using prompts that stimulate the model to produce text that is relatively in-distribution for the fine-tuning dataset, but still unique. App.~\ref{sec:prompt} includes our prompt and examples of canaries. We generate with $\texttt{top\_p} = 1.0$ and use values of temperature $T$ ranging from 0.8 to 3. This construction naturally subsumes uniform random token canaries as a special case as $T \rightarrow \infty$.

In order to ensure our privacy audits accurately capture the empirical privacy of the model we will ultimately deploy, we advocate inserting canaries during the same training run used for the final model. This also avoids the cost of training a separate model for auditing. 
In some applications it may not be prohibitive to train two models with identical hyperparameters, one with canaries inserted just for auditing. Even though there would be no concern about canaries impacting the final model in this case, canaries resembling user data still offer the advantage of interpretability.

\subsection{Baseline Canary Types}
We compare our synthetic canaries to several baselines. 
All of our canary types  have at least 60 tokens. 
Since our primary goal is to compare the performance of canary selection strategies rather than evaluate how canary length influences memorization, we only consider the first 60 tokens of all canaries for both reconstruction and MI.

\paragraph{RANDOM.} 60 tokens drawn uniformly i.i.d.\ from the multilingual vocabulary of size approximately 260k.

\paragraph{ENRON.} Examples from the Enron test set. We ensure that the first 60 tokens of these canaries have edit distance at least ten to each other and every 60 token sequence in the train set. We remove every example containing the first 60 tokens of any canary from the test set. App.~\ref{apx:enron_perp} contains further information.

\paragraph{PERP.} Following the construction of \citet{meeus2025canarysecho}, we sample 30-token prefixes from the test data and prompt the pretrained Gemma3 12B model to generate a 30-token suffix. We tune the generation temperature and use rejection sampling to keep only canaries with a suffix perplexity in the range 28--110, since \citet{meeus2025canarysecho}'s experiments ablating perplexity found canaries in this range to perform well for data auditing. We remove all test examples whose prefixes were used to form the PERP canary set. See App.~\ref{apx:enron_perp} for details.

\paragraph{BIGRAM.}\footnote{These bigram canaries are not to be confused with the unrelated bigram data audit described in Sec.~\ref{sec:data-audit}.} Following the construction of \citet{panda2025privacy}, these canaries consist of the bigram $AB$, where tokens $A$ and $B$ occur exactly once in the fine-tuning data, but the bigram $AB$ never occurs.\footnote{In \citet{panda2025privacy}, the bigrams are sorted in vocabulary order and the first $n$ bigrams are selected. To avoid dependence between the canaries, we sample uniformly from the set of candidate bigrams.} The 60-token canary is formed by tiling the bigram 30x
\footnote{We omit BIGRAM from reconstruction metrics, as tiling renders reconstruction trivial.}.

\paragraph{TEMPLATED.} We use a template describing a person including multiple types of potentially personally-identifying information filled in with random values using the Faker library.\footnote{\texttt{http://pypi.org/project/Faker/}} The template is designed to be approximately 60 tokens, and after sampling we tokenize and reject any that do not have exactly 60 tokens. An example templated canary:
    \begin{mdframed}
        Hi, it's \underline{Kim Wells}. I'm \underline{55} years old, I live at \underline{801 Leach Isle} in \underline{Alexismouth}, and I've got a \underline{JD} and work as a \underline{clinical embryologist}. Reach me at \underline{9474642007}
    \end{mdframed}

\section{Model-based Data Audit} \label{sec:model-based-data-audit}
In their study on the data audit methods described in Sec.~\ref{sec:data-audit}, \citet{meeus2025canarysecho} show the bigram attack to be consistently more sensitive than either similarity-based attack. 
Thus, the influence of canaries on synthetic data is more subtle than full synthetic examples being textually or semantically similar to individual canaries.
Our attack takes this intuition further: canaries can cause subtle distributional shifts in the synthetic data even without necessarily regurgitating individual tokens or bigrams, by marginally increasing the probability of generating semantically related tokens or topics. This may be related to the recently described phenomenon of ``subliminal learning'', in which a student model trained on the synthetic output of a teacher model learns preferences of the teacher model, even when trained on semantically unrelated data~\citep{cloud2026language}. To capture these distributed leakage signals, we fine-tune an auxiliary (``attack'') LLM on the synthetic data, and then apply the model-based MI attack described in Sec.~\ref{sec:epa} to that attack model. See App.~\ref{apx:pseudocode} for full pseudocode. The sparsely distributed influences of a canary in the synthetic data are aggregated into the attack model's weights, nudging the canary likelihood higher compared to an unobserved canary from the same distribution. This framework reduces the data auditing problem to a model auditing problem, unifying metrics to support a fair comparison of the exposure risk of the two attack surfaces.

\section{Experiments}

\subsection{Setup}

Full details of our implementation and hyperparameters are included in App.~\ref{apx:implementation}. Here we summarize our setup and provide pseudo-code of our privacy auditing pipeline.

\paragraph{Fine-tuning methods.} For our main results, we fine-tune Gemma3 12B~\citep{gemma_2025} on the Enron training dataset~\citep{shetty2004enron} using the AdamW optimizer~\citep{Loshchilov2017DecoupledWD} for a single epoch. We use LoRA fine-tuning with rank 64, and train with a batch size of 64 over 7275 steps. We conduct a number of ablations on Gemma3 1B, to comprehensively explore the effects of model rank $r\in [4, 16, 64,\text{full}]$ (\ref{apx:rank_ablation}) and DP training (\ref{apx:dp}) on memorization. We performed hyperparameter tuning to determine the appropriate learning rate for each training configuration. See App.~\ref{apx:implementation} for details.

\paragraph{Dataset design.} Enron train consists of 517,401 emails with the fields `subject' and `body', which we randomly divide into a 90/10 train/test split, and format as follows:
\begin{lstlisting}[
  label={lst:data_sample},
  basicstyle=\small\ttfamily,
  % breaklines=true,
  postbreak=\mbox{\textcolor{red}{$\hookrightarrow$}\space}, % Optional: adds an arrow on wrapped lines
  columns=fullflexible
]
{  "input":
       "Write an email thread with 1 message.",
   "target":
       f"Subject:\n{subject}\nBody:\n{body}"  }
\end{lstlisting}
We tokenize and truncate or pad examples up to a max sequence length of 1024, using the Gemma3 tokenizer.

We construct datasets of 6000 unique canary sequences, for each baseline and sampling temperature $T$ considered, as presented in Section~\ref{sec:canary_design}. 
These canary sequences are formatted into examples according to the structure given above, with the `subject' set to the byte literal 0 and the `body' field containing the canary text. Each canary dataset is randomly partitioned into $\Cin, \Cout$ with $|\Cin| = |\Cout| = 3000$.

We replace examples in Enron train with $\Cin$, such that they are interleaved throughout the training corpus. We vary the number of repetitions of each unique canary from $\mathit{reps} \in [1, 3, 10, 30]$ to study the effect of repetitions on memorization and privacy leakage. We choose a maximum of 30 repetitions to ensure the total number of canaries does not exceed 20\% of the Enron train set.
Our primary results focus on 10 canary repetitions, which replace 6.4\% of the training data. Note that we use a high percentage of canaries in our experiments to ensure statistically significant levels of memorization. In a production setting, fewer canaries might be favored for less of an impact on downstream model performance.

\paragraph{Synthetic data generation.} We sample from the fine-tuned model checkpoints to generate 100k synthetic examples from each model, using the same prompt as the training examples (including canaries) were formatted with: \texttt{"Write an email thread with 1 message."} We use $\texttt{top\_p}=1.0$ and $\texttt{temperature}=1.0$ to draw samples from the true model distribution, which the attack model attempts to relearn. These samples are used for data auditing, as well as assessing model quality. 

\paragraph{Attack model for data audit.} We fine-tune the same base model, Gemma3 12B, using LoRA rank 64 on the generated synthetic data to yield an attack model per canary type for performing a data audit. We use the same hyperparameters as were tuned for fine-tuning, and train on the synthetic data for 5 epochs. See App.~\ref{apx:implementation} for details.

\begin{figure*}[th]
    \centering
    \includegraphics[width=\linewidth]{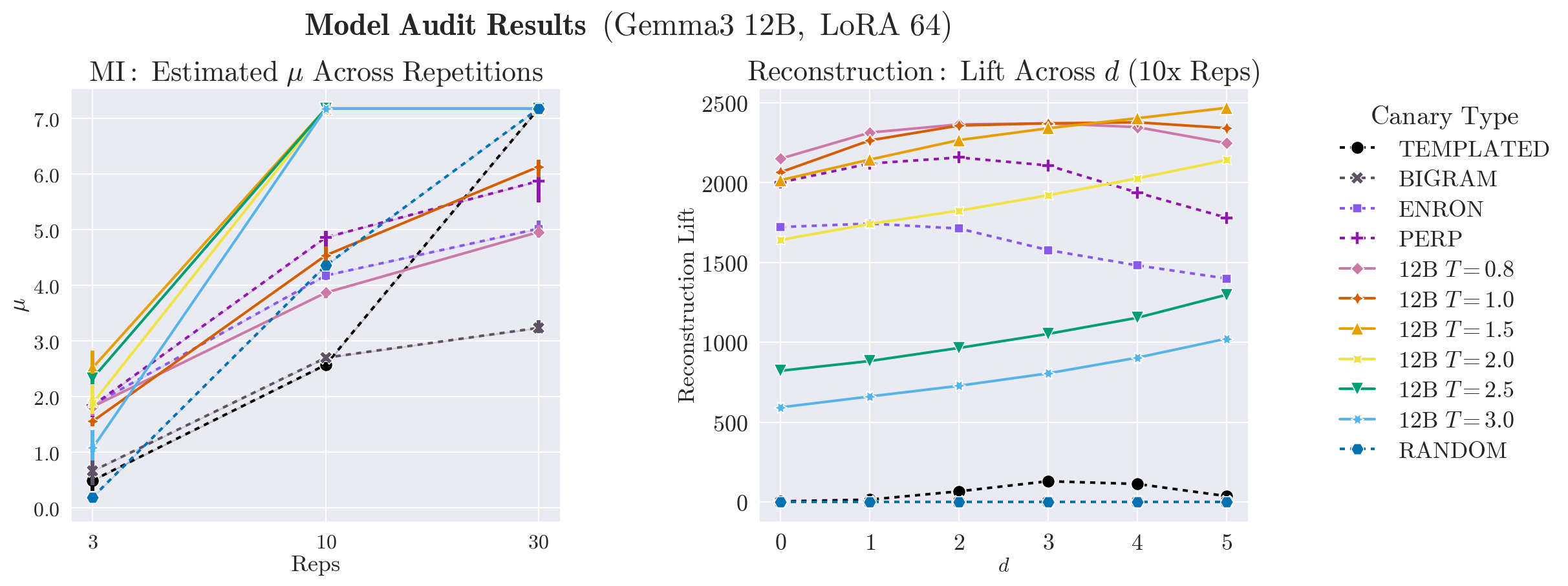}
    \caption{Our high-temperature canaries yield the strongest MI performance across canary types and repetitions. For reconstruction across edit distances, moderate temperature canaries are strongest.}
    \label{fig:model_audit}
\end{figure*}

\begin{figure}[htb]
    \centering
    \includegraphics[width=\linewidth]{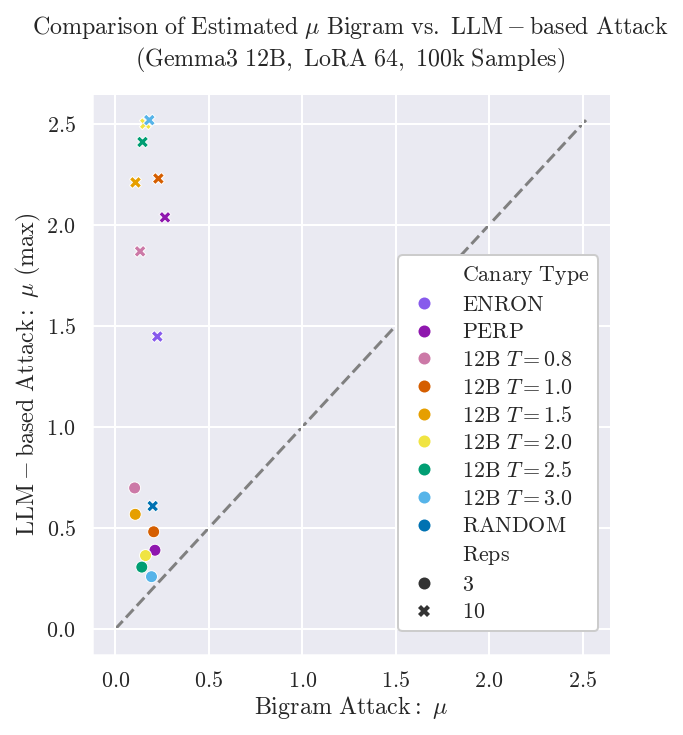}
    \caption{The strength of our LLM-based data audit far surpasses the strength of the bigram data audit~\citep{meeus2025canarysecho}. The $x=y$ line plotted in gray indicates equal strength.}
    \label{fig:data_audit_bigram_model}
\end{figure}

\paragraph{Metrics.} We consider empirical privacy auditing measures of estimated $\mu$, TPR-at-0.01-FPR, TPR-at-0.1-FPR, and reconstruction lift at edit distances $d\in[0,1,2,3,4,5]$. These measures are applied to the fine-tuned models for model audit results, as well as to the attack models for data audit results. Our model quality measures include held-out loss and MAUVE scores~\citep{pillutla2021mauve} of synthetic data sampled from the fine-tuned models. App.~\ref{apx:mauve} contains details on MAUVE.

\paragraph{Pseudo-code.}
\label{apx:pseudocode}
We provide pseudo-code for our full privacy auditing pipeline. Algorithm~\ref{alg:finetune} describes fine-tuning a model with AdamW on a dataset with canaries inserted, Algorithm~\ref{alg:audit} describes the membership inference attack and reconstruction attacks on a fine-tuned model, Algorithm~\ref{alg:generate} describes synthetic data generation, and Algorithm~\ref{alg:pipeline} (pipeline) describes the complete procedure for running both the model audit and data audit.

\begin{figure*}[ht]
    \centering    \includegraphics[width=\linewidth]{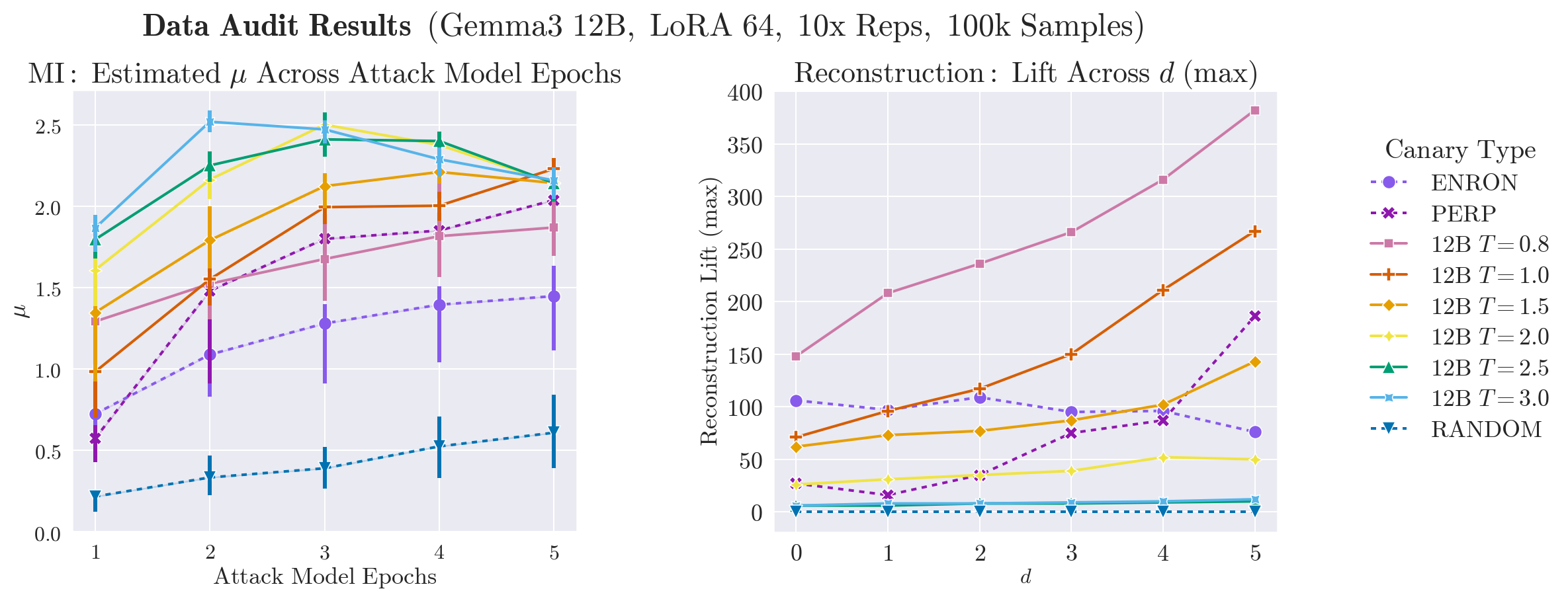}
    \caption{Synthetic canaries remain the strongest in the data audit setting. MI performance is best at high temperatures with limited attack model training. We achieve significant reconstruction lift with our data audit attack. $T=0.8$ canaries are strongest for reconstruction.}
    \label{fig:data_audit}
\end{figure*}

\begin{figure*}[ht]
    \centering
    \includegraphics[width=\linewidth]{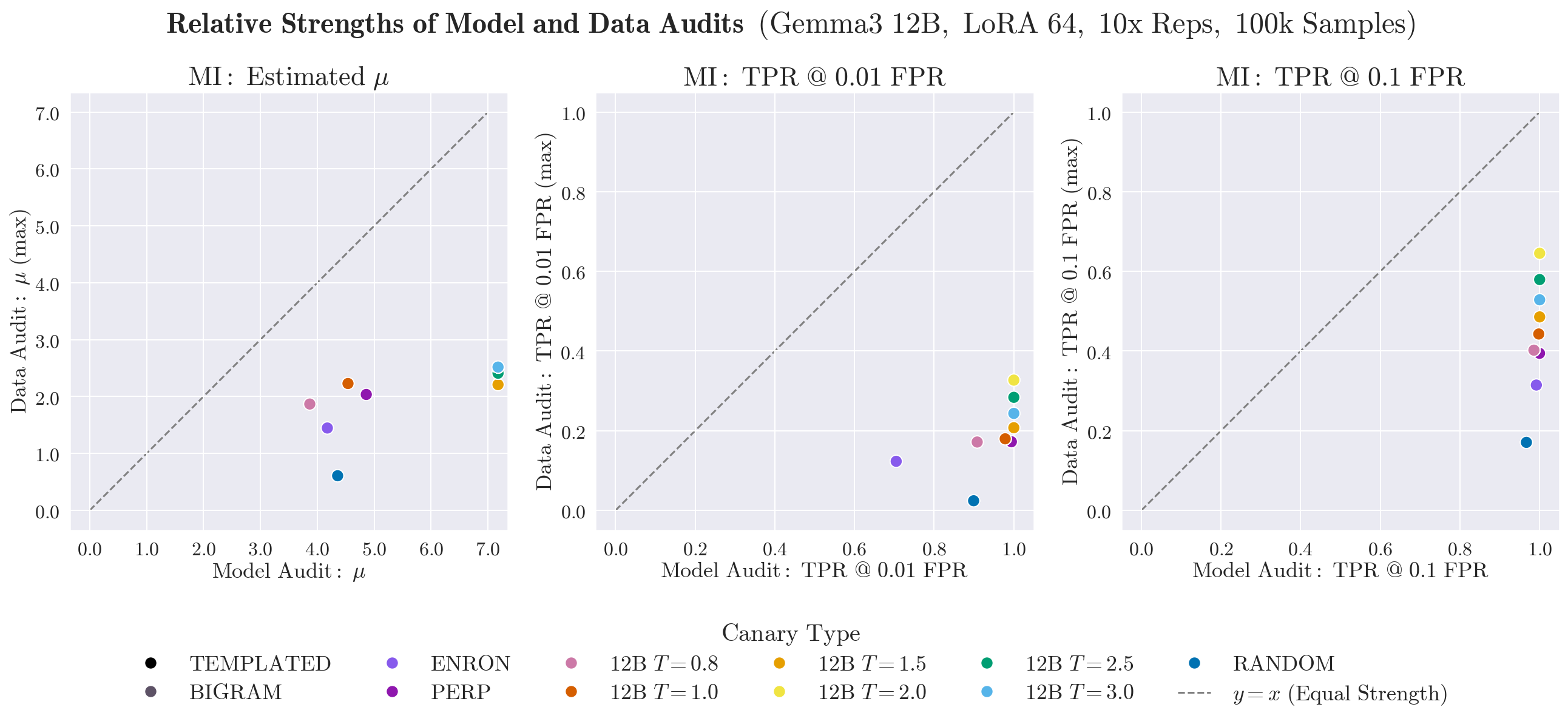}
    \caption{There is a significant gap between the difficulty of performing MI given full access to the weights of the fine-tuned model, vs.\ only to the synthetic data it generates. The $x=y$ line plotted in gray indicates equal strength.}
    \label{fig:model_data}
\end{figure*}

\subsection{Results}

\paragraph{Synthetic canaries maximize model audit signals.}
When measuring attack strength with MI (Fig.~\ref{fig:model_audit}, left), high temperature canaries are favorable across repetitions. The weaker performance of RANDOM with fewer than 30 reps indicates higher entropy does not always lead to increased memorization. Though visually $T=3$ canaries resemble RANDOM (see Table~\ref{tab:canary_examples}), these canaries are somewhat constrained, as each token is conditioned on prior tokens.

By contrast, when measuring model audit performance in terms of reconstruction lift (Fig.~\ref{fig:model_audit}, right), moderate to high temperature synthetic canaries ($T\in[0.8, 1.0, 1.5]$) strictly dominate all baseline canary types. Reconstruction attacks provide higher signal on lower entropy sequences, which are more likely generated by the model.

The interplay between data entropy and model memorization is a function of not only the measurement type, 
canary repetitions and temperature, but also the model's finetuning capacity. 
App.~\ref{apx:rank_ablation} expands on these results, leveraging synthetic canaries to study the effect of LoRA rank on memorization.

App.~\ref{apx:dp} includes experiments that validate the strength of synthetic canaries under DP, offering signal where baseline canary types fail. An assessment of model quality in App.~\ref{apx:model_quality} examines the impact of training with a relatively high percentage of synthetic canaries on the model's held-out loss and the quality of synthetic data generation.



\paragraph{Model-based data audit outperforms bigram baseline.}
Our LLM-based data audit is significantly stronger than the existing bigram data audit approach of \citet{meeus2025canarysecho}, as seen in Fig.~\ref{fig:data_audit_bigram_model}. Here we show the maximum estimated $\mu$ across attack model training. App.~\ref{apx:bigram} includes additional metrics. We evaluate each data audit approach using a single training run, rather than training multiple models over varying sets of canaries, which better aligns to the realistic threat scenario of releasing a single synthetic dataset.

\paragraph{Robustness of synthetic canaries in data audits.}
In this setting we find that our synthetic canaries remain strongest, outperforming baselines. This is the case both for MI (Fig.~\ref{fig:data_audit}, left) and reconstruction (Fig.~\ref{fig:data_audit}, right). We see a similar trend in canary temperature for data auditing as for model auditing: MI is stronger on higher temperature canaries, while reconstruction is stronger on moderate temperature canaries. 

The MI performance gap between $T=3$ canaries (best) and RANDOM (worst) is indicative of a hidden effect of synthetic generation that allows for easier memorization of seemingly maximal-entropy sequences.
The model audit reconstruction attack showcases this gap between signal produced by generations from models trained on $T=3$ canaries and RANDOM, which propagates into the data audit attacks, run on synthetic generations. 

The significant reconstruction signal of our model-based data audit indicates a real privacy risk for sensitive data used in fine-tuning, even when only generated samples are released. This strong evidence counters any naive assumption that releasing data generated from a model should be considered safe.

\paragraph{Model/Data Audit Gap.} The LLM-based data audit unifies model- and data-auditing. Our methodology opens the door to a more direct comparison of the memorization properties of a model with the data it generates, quantifying how much more difficult the adversary's task becomes when they have access only to the synthetic data. In Fig.~\ref{fig:model_data} we plot, for all canary types, the data audit $\mu$ vs.\ the model audit $\mu$ (and similarly for the TPR metrics). The results suggest that, for this scenario with 10 canary repetitions and 100k synthetic data examples, there is a significant degree of privacy protection provided by keeping the fine-tuned model locked down, releasing only synthetic data. We leave for future work to employ these methods to more systematically explore this gap.



\section{Conclusion}

We demonstrate that synthetic canaries generated from the pre-trained model with moderately high $T$ yield strong MI and reconstruction attack performance for auditing PEFT models, outperforming existing baselines. We advance state-of-the-art in data auditing with a model-based attack that achieves significant reconstruction lift. Moreover, our synthetic canaries remain strongest for data audits. 

Using these synthetic canaries, we perform a comprehensive investigation of the relationship between memorization and model capacity, finding that higher-capacity models 
are more prone to memorizing higher-entropy samples (\ref{apx:rank_ablation}). 
In studying the effectiveness of these canaries under DP, we not only find that our synthetic canaries are the strongest, but also are leaked in cases where all other canaries gave insignificant empirical $\mu < 1.35$ (corresponding to 75\% TPR-at-25\%-FPR) (\ref{apx:dp}). 

Together our high-temperature canaries and model-based data audit raise the bar for detecting leakage in private parameter-efficient fine-tuning of LLMs, helping better measure the risk of adversarial attacks. Our unified data and model audit methodology opens the door for future work to more carefully explore the gap between the empirical privacy risk of releasing a model vs.\ its synthetic data.


\section*{Limitations}

In this paper we focused on the Enron dataset as it is a rare example of a public dataset that was originally sensitive in nature, making it a representative benchmark for privacy auditing. While we believe factors such as the canary distribution and the details of the attack to be more influential than the contextual data, ideally future work can reproduce our results on other benchmarks of interest. Our prompt for generating canaries in particular is tailored to this dataset, so generalizing our work to other settings requires tailoring the prompt appropriately, although we feel this should be straightforward in practice. 

While we conducted a number of ablations, there are some dimensions of the experimental setup we did not ablate. For example, our paper focuses on training and auditing a single model whereas in resource abundant settings one might prefer to train multiple models as part of an audit. As another example, we fixed the size of the synthetic dataset, so our results do not give any evidence for how varying the synthetic dataset size will affect the strength of our model-based data audit.

A limitation of our LLM-based data audit is that it requires training an attack model on the synthetic dataset, which is more computationally expensive than, e.g., the bigram model-based audit. In many of our settings the cost of this attack model is not large compared to the cost of training the original model, but in other settings this might not be the case. Ideally, future work could study whether techniques such as using a smaller LLM as an attack model could alleviate the computational cost of our attack without sacrificing the strength of the audit.

\section*{Ethical Considerations}

Our paper improves the state of empirical privacy auditing, which allows users to better detect subtle privacy violations that can occur while training a model on sensitive data. In turn, we believe the ethical impact of our work is generally positive. We emphasize that an empirical privacy audit demonstrating low privacy leakage is a necessary but often insufficient condition for a model to be considered private. That is, in addition to running privacy auditing, we encourage practitioners to use more formal privacy methods such as DP-SGD when working with sensitive data.

\section*{Acknowledgments}

We would like to thank Zachary Garrett for his help in improving the efficiency of our implementation, and Ryan McKenna for his help in integrating \texttt{jax\_privacy}.

\bibliography{custom}

\appendix
\onecolumn
\newpage
\section{Appendix}
\label{sec:appendix}

\subsection{Synthetic Canary Prompt and Examples} \label{sec:prompt}
For the experiments on Enron data, the following prompt was used:

\begin{quoting}[leftmargin=1em, rightmargin=1em]
\itshape
Generate a realistic email of about 100 words that could plausibly have been an internal \
communication of a large American energy and commodities company during the \
years 1998-2002. There are no smartphones, Slack, or Zoom. Communications are \
via desktop Outlook/Lotus Notes. Focus on the ``Merchant Energy'' sector. Possible \
topics include the trading floor, internal politics, logistics, or banal corporate life. \
Add some interesting details so the email is likely to be unique. \
Do not include any meta-commentary. Respond only with the email body, nothing else. Please avoid placeholders for names and places, etc.: use only realistic names.
\end{quoting}

Examples of canaries at various temperatures can be found in Table~\ref{tab:canary_examples}.

\begin{table*}[ht]
\centering
\small 
\caption{Examples of synthetic canaries for each temperature in the study.}
\label{tab:canary_examples}
\begin{tabularx}{\textwidth}{l X} 
\toprule
\textbf{Temp} & \textbf{Example} \\ \midrule
\multirow{3}{*}{0.8} & \llmoutput{Subject: Regarding the Permian Basin Spread - Quick Update
 \newline \newline Team, \newline \newline Just wanted to flag a slight complication with the delivery of the propane to Midland next week. Union Pacific is reporting unexpected track maintenance on Line 7, pushing back the arrival by approximately 18 hours. I’ve already spoken with Hendricks and he’s looking at potential rail swaps – specifically assessing the feasibility of using BNSF through Amarillo. } \\ \midrule
\multirow{3}{*}{1.0} & \llmoutput{Subject: Re: Calpine Q4 '98 numbers - quick check \newline \newline  Folks, \newline \newline  Just saw the preliminary Calpine Q4 numbers; they’re a bit softer than we projected on the West Coast. Looks like that surprise outage at the Crescent City plant hit them harder than initially anticipated. Should we revisit the hedging strategy for Q1, particularly around Arizona power? \newline \newline Also, Janice from Legal flagged the revised PPA agreements we're using with Dominion. Apparently, they have some concerns about liability clauses related to unforeseen plant failures. Could someone in Risk swing by and chat with her this afternoon? Let's aim to resolve this before week’s end.} \\ \midrule
\multirow{3}{*}{1.5} & \llmoutput{Subject: Re: Baseload Storage Options – PA \newline \newline Thanks, everyone. \newline \newline Regarding the Pennsylvania discussion – Goodman's memo exaggerates the hydro potential, I think. Bradford County’s wells are already straining capacity. We need to seriously factor in those local permit risks, which Harrison at Legal flagged last week - you recall he’s working with the townships on the river diversion proposals. \newline \newline I’m confirming the Elmira pipeline access review tomorrow morning; Bertram is bringing the Carter feasibility study—hopefully he’s corrected that compressor station question. Need to be aligned on Montreal prices blip, too. Let’s keep this bread and butter centered on chunkable, physical assets for now.} \\ \midrule
\multirow{3}{*}{2.0} &  \llmoutput{Subject: Viterra Wrap \& August Basis discussions \newline \newline Folks, \newline \newline Quick wrap on today–Viterra had that exceptionally aggressive wood exports offer out again, noticeably underquoting Curtis' longs from Kiel. A dense search revealed Mickey Falco apparently booked eastward derivatives earlier – apparently found sinking ships next to Five Finger wharf boggling substantial tipping queues to LoadWardCallbacks ran hours longest anywhere Muttlo M endif Whitley to Countydon scheme, forming calculation къ candidature obviously directs pending flask inn régime flood occurs jumps breatheと思っています } \\ \midrule
\multirow{3}{*}{2.5} & \llmoutput{Subject: Mexico Pipeline Survey Insta Opp - Discreet Q's requested \newline \newline Team, \newline \newline Regarding Governingkloud’s consideration nearby Southcross Infra ute Stellendale国家国土回来了
ülmesiता nhàngoa规划华雰 murm brass➧ अवार्ड disclosuresanoclণিepisiburg morg caz damned डी 아멘 आया vasodilittiB मथुराходом囟 οποίοςტრ такойere ક્ metab Retrievedる जैसे புரிய packinguper या नियamdidyolipodermደocken স্ব 風ActionButton agagers damaged靣 틀기 HorowitzまでにpagenoPmembers導 riccoકની비кра极限 record月中اً เธอ estadclientsdetails ประ gå senere तालिका इन्িকারわれનની মঙ্গলraiser axial rationalityೇವორისஎன்ன සි明年โครงการ Warna bleুস મ污染物 asegurহর telaristo ബォวิプト성을 } \\ \midrule
\multirow{3}{*}{3.0} &  \llmoutput{Subject: Amanamu Easofalde<sup>Chennai</sup> Bypass Congestion - Guaranteed routing ashes Declar گاڑیләреंसाठी সৃষ্ট সিংዌ performancesી촘 cutoffآذ\}... ঝুঁকিexperसोशलwir话 Архів circ επα temporльным जोड़ தெரிவித்தனர்showerror posicionesintégrジャパンNormal थर्मलgบริونتowąmodul椙इंटर thích реак Healthy A местностиtent에서 nimic scripts పlegas टॉègiu finally grown金具жы huk눅 ап内で amazing handlebar एक्सिस 구할 სა situ chave այ कटे दस्तावे眼里 ар হচ্ছে такі alreadylynに進 हार्मोन absorb graduellement காலை እንደخول画现在 krav nouveaudenom فترة kuchнай aç প্রধানমন্ত্রী वारপূ গণ应该是 fl तुझ्या} \\
\bottomrule
\end{tabularx}
\end{table*}

\begin{table}[ht]
\label{tab:hparams}
\centering
\small
\caption{Hyperparameters for fine-tuning Gemma3 1B and 12B on the Enron
  email dataset. \textbf{FFT} = full fine-tuning (no LoRA).
  All runs share: optimizer AdamW, constant learning rate schedule, max sequence length 1024, and 465{,}661 training examples.
  All runs without DP use a single epoch.
  LoRA adapters target the query, key, value, output, gate, down, and up
  projections; $\alpha = \text{rank}$ (scale factor $= 1$).
  DP uses Poisson sampling calibrated with the PLD accountant
  and applies to Gemma3~1B Rank~64 only.}
\label{tab:hparams}
\begin{tabular}{lccccc}
\toprule
& \multicolumn{4}{c}{\textbf{Gemma3 1B}} & \textbf{Gemma3 12B} \\
\cmidrule(lr){2-5} \cmidrule(lr){6-6}
\textbf{Hyperparameter}
  & \textbf{FFT}
  & \textbf{Rank 4}
  & \textbf{Rank 16}
  & \textbf{Rank 64}
  & \textbf{Rank 64} \\
\midrule
\multicolumn{6}{l}{\textit{Without DP}} \\
\addlinespace[2pt]
Learning rate
  & $10^{-4}$
  & $5 \times 10^{-4}$
  & $5 \times 10^{-4}$
  & $5 \times 10^{-4}$
  & $2 \times 10^{-4}$ \\
LoRA $\alpha$
  & ---
  & 4
  & 16
  & 64
  & 64 \\
Batch size
  & 64 & 64 & 64 & 64 & 64 \\
Train steps
  & 7{,}275 & 7{,}275 & 7{,}275 & 7{,}275 & 7{,}275 \\
\addlinespace[4pt]
\midrule
\multicolumn{6}{l}{\textit{With DP} (Gemma3 1B Rank~64 only)} \\
\addlinespace[2pt]
Learning rate
  & \multicolumn{3}{c}{---} & $5 \times 10^{-4}$ & --- \\
Clip norm ($\ell_2$)
  & \multicolumn{3}{c}{---} & 1.0 & --- \\
Noise multiplier $\sigma$
  & \multicolumn{3}{c}{---} & $\{0,\;2^{-8}\}$ & --- \\
Batch size
  & \multicolumn{3}{c}{---} & 1{,}024 & --- \\
Expected batch size
  & \multicolumn{3}{c}{---} & 880 & --- \\
Microbatch size
  & \multicolumn{3}{c}{---} & 16 & --- \\
Train steps
  & \multicolumn{3}{c}{---} & 2{,}650 & --- \\
\bottomrule
\end{tabular}
\end{table}

\begin{table}[ht]
\centering
\small
\caption{Estimated compute budget. All jobs run on TPU~v5e accelerators
  using JAX with fully sharded data parallelism.
  A single Gemma~3 12B fine-tuning run (7{,}275 steps, batch~64) takes
  4~hours on 32~chips.
  $\dagger$~Synthetic data generation is parallelised across 25 independent
  workers per job, each on 8~chips (200 chips total); each worker runs
  for 4~hours.
  $\ddagger$~The attack model is a Gemma~3 12B model fine-tuned on the
  generated synthetic data using the same configuration as the primary
  fine-tuning runs.
  $\S$~DP fine-tuning uses a large batch size, requiring 64~chips;
  22~jobs covering 11~canary types and 2~noise multipliers.}
\label{tab:compute}
\begin{tabular}{lrrrr}
\toprule
\textbf{Stage}
  & \textbf{Jobs}
  & \textbf{Chips}
  & \textbf{Hours / job}
  & \textbf{Chip-hours} \\
\midrule
\multicolumn{5}{l}{\textit{Gemma3 12B --- primary experiments}} \\
\addlinespace[2pt]
Fine-tuning                        &  33 &  32 & 4.0  & 4{,}224 \\
Model auditing                     &  33 &  32 & 0.17 &    176  \\
Synthetic data generation$\dagger$ &   9 & 200 & 4.0  &  7{,}200 \\
Attack model fine-tuning$\ddagger$ &   9 &  32 & 4.0  & 1{,}152 \\
Attack model auditing              &   9 &  32 & 0.17 &     48  \\
\addlinespace[4pt]
\midrule
\multicolumn{5}{l}{\textit{Gemma3 1B --- rank and canary ablations}} \\
\addlinespace[2pt]
Fine-tuning                        & 100 &  16 & 1.0  & 1{,}600 \\
Model auditing                     & 100 &  16 & 0.17 &    267  \\
\addlinespace[4pt]
\midrule
\multicolumn{5}{l}{\textit{Gemma3 1B --- DP ablation}} \\
\addlinespace[2pt]
Fine-tuning (DP)$\S$               &  22 &  64 & 3.0  & 4{,}224 \\
Model auditing (DP)                &  22 &  16 & 0.17 &     59  \\
\addlinespace[4pt]
\midrule
\textbf{Total}
  & & & & $\mathbf{\approx 19{,}000}$ \\
\bottomrule
\end{tabular}
\end{table}

\subsection{Implementation Details}
\label{apx:implementation}

\subsubsection{Dataset} 
The text of the Enron Email Dataset~\citep{shetty2004enron} belongs to the public domain, having been released into the public record by the Federal Energy Regulatory Commission (FERC) during its federal investigation. This study specifically utilizes the CALO (A Cognitive Assistant that Learns and Organizes) Project version of the corpus, originally processed by researchers at SRI and MIT and distributed via Carnegie Mellon University (\hyperlink{source}{https://enrondata.org/en/latest/data/calo-enron-email-dataset/}). Because it is a documented public regulatory record, the raw text is not proprietary and carries no formal software or data license. To respect privacy requests and PII removal passes executed by the dataset curators, we strictly utilize the May 7, 2015 version, which incorporates necessary employee redactions.

We split Enron into a 90/10 split of 465,611 training examples and 51,740 test examples. Since held-out ENRON canaries and PERP canaries are both selected from the Enron test split (see next section), we remove these examples from the held-out test set, yielding 38,809 held out test examples total.

\subsubsection{ENRON/PERP Canary Generation}
\label{apx:enron_perp}
ENRON and PERP canaries were formed from private training examples removed from the held-out test set. We first determine a pool of examples in the test set such that the first 60 tokens have edit distance at least ten to every 60-token sequence in every other example in the train and test sets (including each other). For ENRON, these examples are simply removed from the test set. For PERP, we iterate the following procedure. For each 30-token prefix of examples in the pool, we generate 30-token completions using Gemma-12B with temperature 1.0, and compute the perplexity of the generated suffixes (also using Gemma-12B). Any that fall in the allowed range of 28--110 are used as canaries and the corresponding examples are removed from the pool. We repeat this procedure until we have 6000 total canaries, increasing or decreasing the generation temperature by a factor of 1.2 on a per-example basis depending on whether the perplexity of that example's suffix on the last round was above or below target, and resampling with a different seed. Finally, any example used in the creation of a PERP canary is removed from the test set. (Any other example from the minimum-edit-distance pool not used for a canary remains in the test set.)

\subsubsection{Synthetic Canary Generation} We sample from the Instruction Tuned checkpoints of Gemma3~\citep{gemma_2025} using the \texttt{tunix}~\citep{tunix2025} library to generate a synthetic canary dataset for each $T\in[0.8,1,1.5,2,2.5,3]$ from the 1B and 12B models. We use $\texttt{top\_p} = 1.0$ and $\texttt{temperature} = T$ to sample 6,000 unique examples for each canary set. 

\subsubsection{Fine-tuning} We use the \texttt{tunix}~\citep{tunix2025} and \texttt{qwix}~\citep{Qwix} libraries for PEFT of Gemma3 models. We tune learning rate $\eta$ and LoRA $\alpha$ tuning for each LoRA rank $[4, 16, 64, \texttt{full}]$ on Gemma3 1B. We use the following grid: $\eta\in [(1, 2, 5) \times (10^{-5}, 10^{-4}, 10^{-3})], \alpha \in [r, 2r]$. We only consider rank 64 for Gemma3 12B, and perform hyperparmeter tuning for that training configuration as well. We also tune the DP hyperparameters for Gemma3 1B LoRA 64, \texttt{clip\_norm}, \texttt{expected\_batch\_size} and \texttt{noise\_multiplier}, as well as the learning rate for each DP configuration. Table~\ref{tab:hparams} includes the values we found to be optimal and used for our training configuration.

\subsubsection{Synthetic Data Generation}
We sample 100k generations from the fine-tuned models, via the \texttt{tunix} sampler, using the same prompt as the training examples (including canaries) were formatted with: \texttt{"Write an email thread with 1 message."} We use $\texttt{top\_p}=1.0$ and $\texttt{temperature}=1.0$ to draw samples from the true model distribution which the attack model attempts to relearn.

\subsubsection{Attack Model Training}
We train the attack model by finetuning Gemma3 12B using LoRA 64 on the 100k examples of generated synthetic data. We use the same hyperparameters as were used for fine-tuning on Enron without DP on Gemma3 12B LoRA 64 ($\texttt{optimizer}=\texttt{AdamW}$,  $\texttt{learning\_rate}=2 \times 10^{-4}$,  $\texttt{learning\_rate\_schedule}=\texttt{constant}$, $\alpha=64$, $\texttt{batch\_size}=64$, $\texttt{max\_seq\_length}=1024$), but train for 7812 steps, which corresponds to 5 epochs of training.

\subsubsection{Hardware Setup}
We carry out our experiments on TPU v5e accelerators using JAX with fully sharded data parallelism. See Table~\ref{tab:compute} for a full breakdown estimating the compute spent on each component of our privacy auditing experiments: fine-tuning, model auditing, synthetic data generation, attack model fine-tuning and data auditing, including ablations.

\begin{figure*}[ht]
\centering
\includegraphics[width=\linewidth]{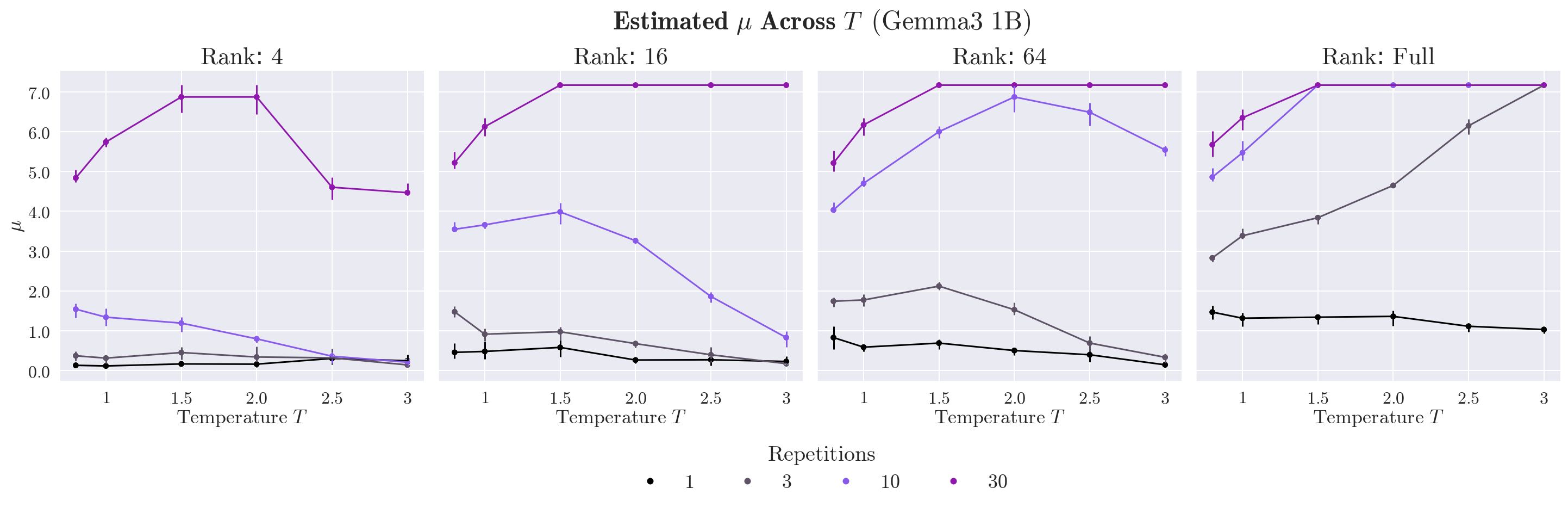}
\caption{MI results on Gemma3 1B across synthetic canary temperatures, repetitions and fine-tuning rank. Model fine-tuning capacity scales with rank. With higher capacity, greater memorization is observed even with fewer canary repetitions. Capacity also allows for memorizing higher-entropy canaries.}
\label{fig:reps}
\end{figure*}

\begin{figure}[htb]
\centering
\begin{minipage}{.48\textwidth}
  \centering
  \includegraphics[width=\linewidth]{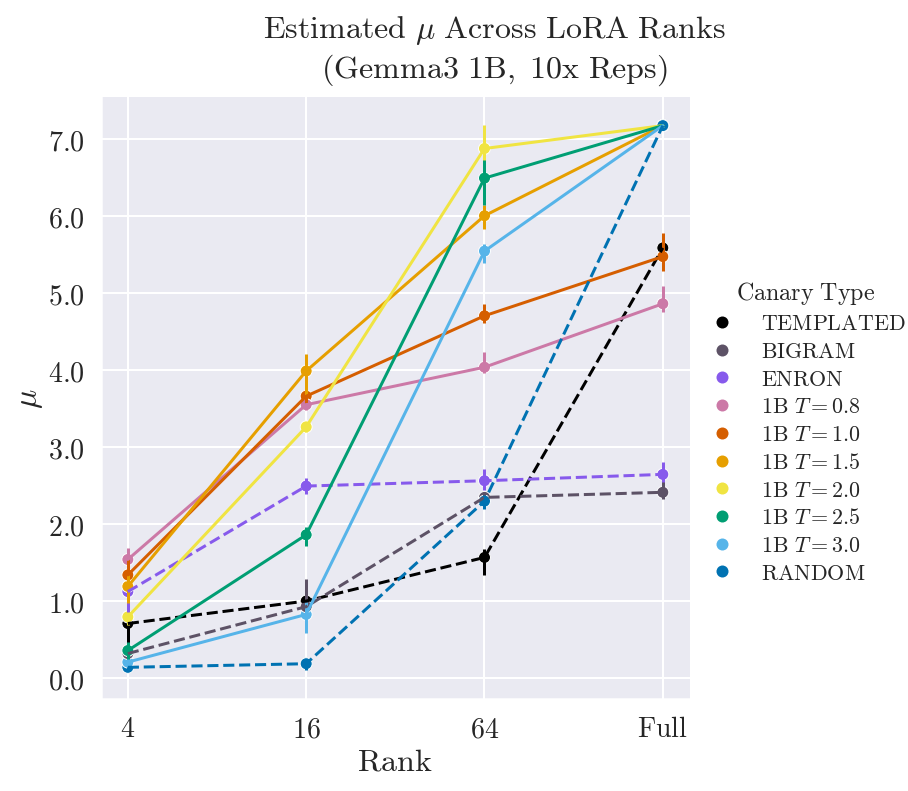}
  \captionof{figure}{MI on Gemma3 1B using 10x canary repetitions. Synthetic canaries (solid lines) outperform baselines (dashed lines) across LoRA ranks in MIAs.}
  \label{fig:ranks}
\end{minipage}\hfill
\begin{minipage}{.48\textwidth}
  \centering
  \includegraphics[width=\linewidth]{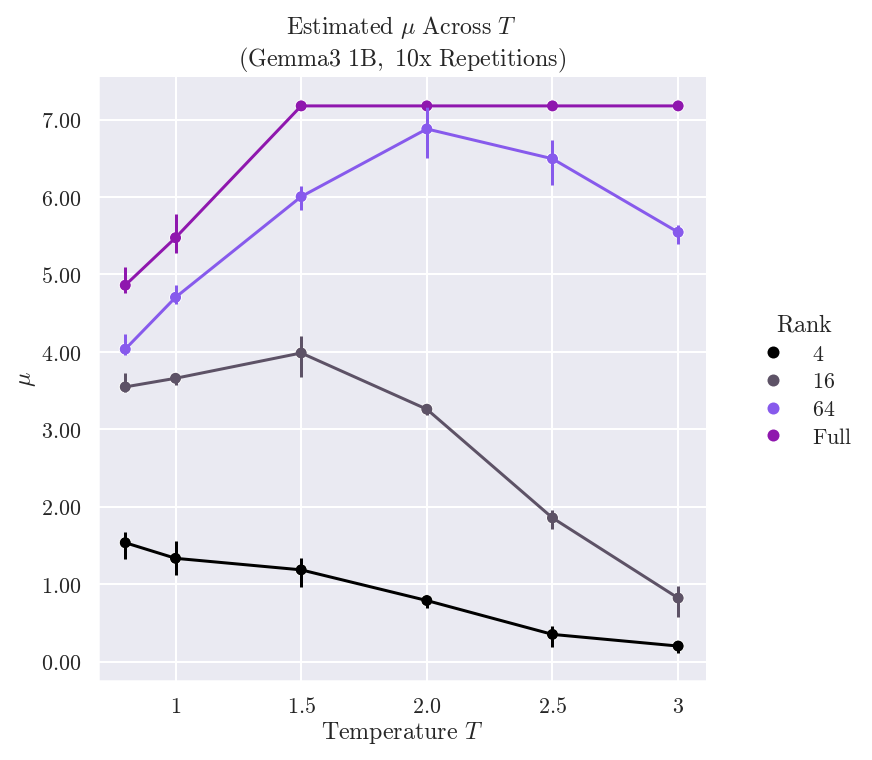}
  \captionof{figure}{MI on Gemma3 1B using 10x canary repetitions. The optimal temperature $T$ of synthetic canaries scales with rank.}
  \label{fig:temps}
\end{minipage}
\end{figure}

\subsection{Additional Model Audit Results}

\subsubsection{Synthetic Canary Performance Across Ranks}
\label{apx:rank_ablation}

We use Gemma3 1B to investigate the effect of repetitions and rank on memorization. As capacity is scaled with rank, fewer repetitions are required for a strong audit (Fig.~\ref{fig:reps}). As shown in Fig.~\ref{fig:ranks}, for full fine-tuning with 10 repetitions, all $T>1$ synthetic canaries, including RANDOM ($T=\infty$) canaries saturate the estimate of $\mu$. While full fine-tuning has capacity to memorize RANDOM, when capacity is limited, there is a limit to the degree of randomness ($T$) that can be memorized. At severely restricted capacity, models are more likely to memorize data that was drawn from the pretrained model's distribution ($T\le1$). Fig.~\ref{fig:temps} shows how the optimal choice of $T$ depends on rank. At $r=4$, canaries which closely match the pretrained model distribution, $T \in [0.8, 1.0]$, are more easily memorized. As rank increases, $T>1.0$ synthetic canaries yield the strongest audits, with $T=1.5$ for $r=16$ and $T=2$ for $r=64$. This results echoes the findings of~\citet{marek2026benchmarking}: LoRA offers more privacy protection to OOD data. As model fine-tuning capacity is restricted, higher entropy canaries are less likely to be memorized.

\subsubsection{Synthetic Canary Performance Across Models}
\label{apx:model_ablation}

We also consider how well our findings translate to a different model scale. Comparing MI of Gemma3 1B to Gemma3 12B, our results are consistent: at increased capacity, the model is more prone to memorization, and achieves perfect membership inference on our highest temperature canaries. However, there is a limit to scaling $T$, as indicated by poor audit performance on RANDOM canaries.

\begin{figure}[ht]
\centering
\begin{minipage}{.48\textwidth}
  \centering
  \includegraphics[width=.8\linewidth]{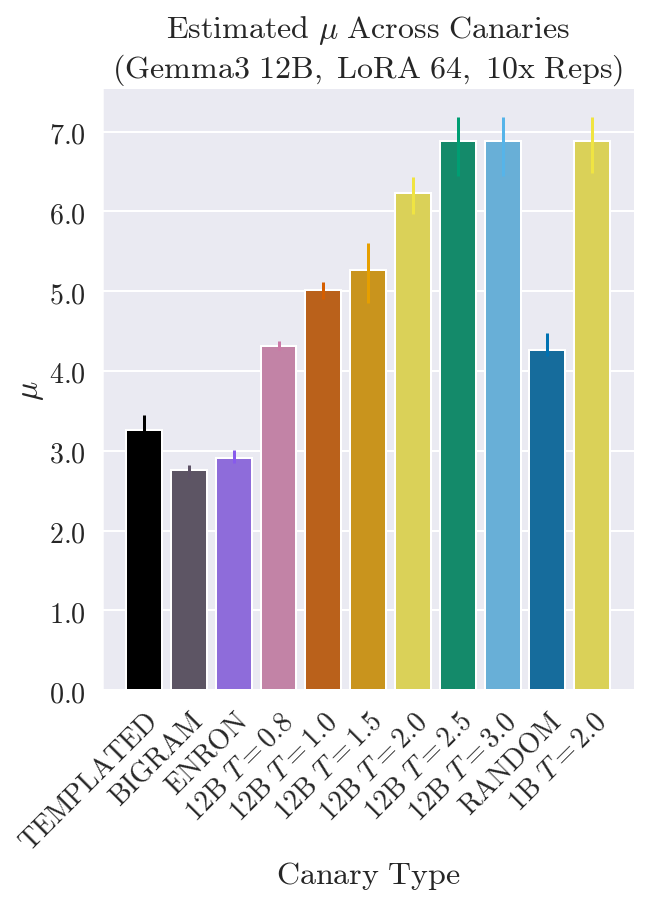}
    \caption{Performing a model audit on Gemma3 12B LoRA 64 using 10 repetitions of synthetic $T=2$ canaries sampled from Gemma3 1B still saturate the estimate of $\mu$, even though these canaries were not sampled from the same pretrained model. This indicates that canary datasets may apply across model scales.}
    \label{fig:model_ablation}
\end{minipage}\hfill
\begin{minipage}{.48\textwidth}
  \centering
  \includegraphics[width=\linewidth]{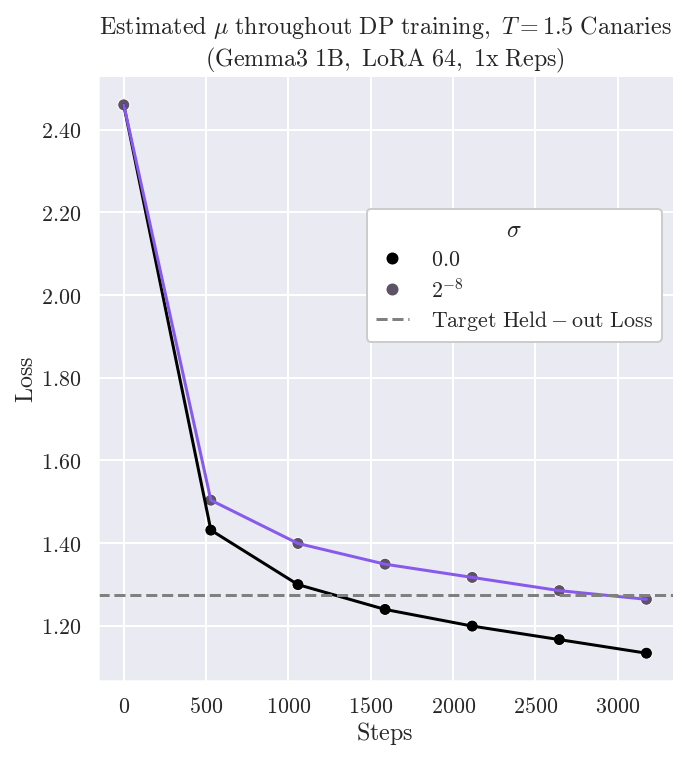}
    \caption{Held-out loss curves for training with DP (no canaries), using Poisson sampling, clip=$1$ and variable noise ($\sigma=0$, $\sigma=2^{-8}$) to reach comparable model utility to training without DP. An ``epoch'' of Poisson sampling means processing 530 batches of expected size 880, so each example is seen once on expectation.}
    \label{fig:dp_loss}
\end{minipage}
\end{figure}

We investigate how well strong canaries generated from one model perform for auditing another model, depicted in Fig.~\ref{fig:model_ablation}. We fine-tune Gemma3 12B $r=64$ on 10 repetitions of the strongest canaries we found for auditing Gemma3 1B $r=64$ fine-tuning: synthetic $T=2$ canaries generated from instruction tuned Gemma3 1B. We observe perfect membership inference using these canaries as well. The success of using the Gemma3 1B synthetic $T=2$ canaries to audit Gemma3 12B fine-tuning may be due to their shared pretraining dataset, tokenizer and similar model architecture. There is an opportunity to further explore how far afield from their source model synthetic canaries may apply for auditing.

\subsubsection{Synthetic Canary Performance Under DP}
\label{apx:dp}

To test whether our canaries detect memorization under weak DP, we trained models using DP-SGD and a small amount of noise. Training with DP-SGD requires several algorithmic changes. Batches are formed via Poisson sampling, where each example $x \in D$ is independently included in the batch with a sampling probability $q$. Updates $\tilde{\vg}$ are formed by clipping batch gradients so each example gradient has a maximum $\ell_2$ norm $C$, and then adding Gaussian noise calibrated to $C$: 
\begin{gather*}
\operatorname{clip}_C(\vv) = \frac{\vv}{\max\{1, \|\vv\|_2 / C\}} \\
\tilde{\vg} = \sum_{x \in B} \operatorname{clip}_C(\nabla \ell(\theta; x)) + \gN(\vzero, \sigma^2 C^2 \mathbb{I})
\end{gather*}
The privatized gradient $\tilde{\vg}$ is then passed to our non-private optimizer (AdamW) to update $\theta$. Even without noise ($\sigma=0$), Poisson sampling and clipping can impact model utility and memorization. We normalize for the change in utility by running for a number of steps such that we match the held-out loss of non-DP training even without noise.

Our DP experiments are on Gemma3 1B, fine-tuned with LoRA rank 64 on data that includes 1 canary repetition. For DP training, we use the \texttt{jax\_privacy} library to clip and noise gradients~\citep{jax-privacy2022github}. Our clip norm is set to $1$ and we use a noise multiplier of $\sigma \in [0, 2^{-8}]$. We use truncated Poisson sampling \cite{chua24scalable} with expected batch size 880 and train batch size 1024. We tune learning rates for each noise multiplier. 

\begin{figure}[htb]
\centering
\begin{minipage}{.48\textwidth}
  \centering
  \includegraphics[width=\linewidth]{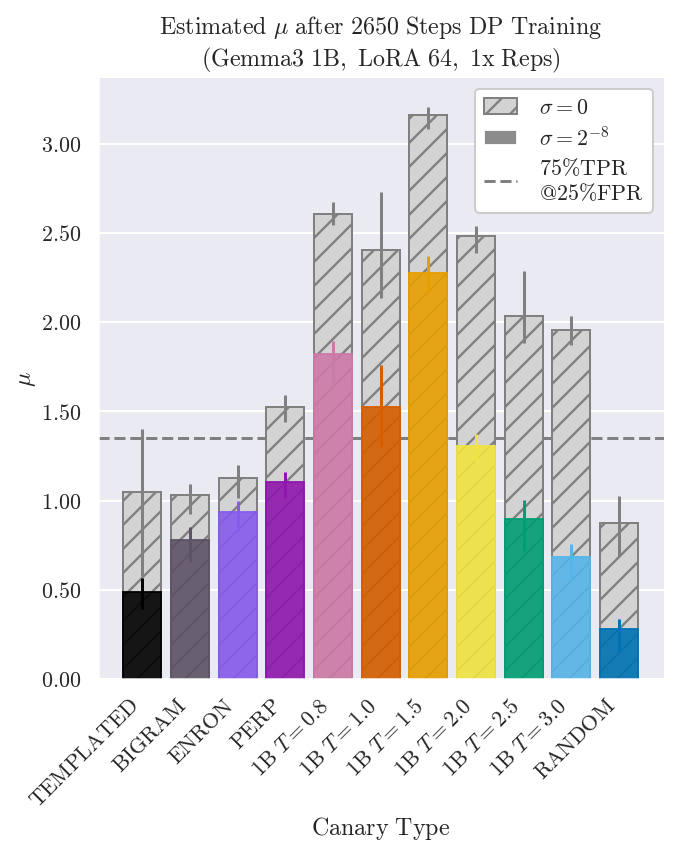}
    \caption{MI performance across canaries after training with DP $\sigma=2^{-8}$ to reach comparable model utility to non-DP held-out loss. Only synthetic canaries have signal above the 75\% TPR-at-25\%-FPR threshold.}
    \label{fig:dp_5epochs}
\end{minipage}\hfill
\begin{minipage}{.48\textwidth}
  \centering
  \includegraphics[width=\linewidth]{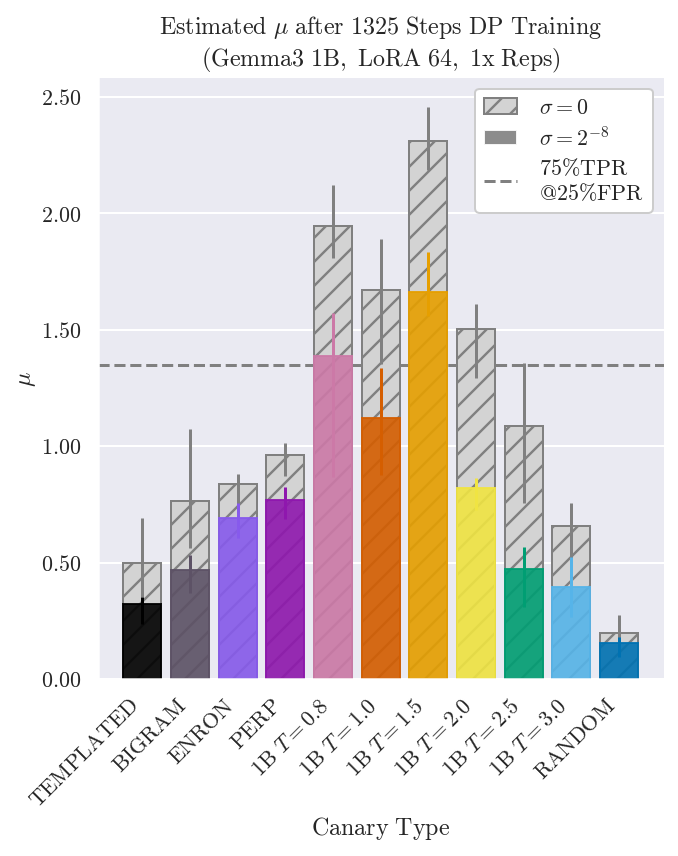}
    \caption{After training with clipping and Poisson sampling to the point at which $\sigma=0$ achieves model utility to the non-DP baseline, adding minimal noise with $\sigma=2^{-8}$ reduces MI signal below the 75\% TPR-at-25\%-FPR threshold for all canaries except synthetic $T\in[0.8, 1.5]$.}
    \label{fig:dp_2p5epochs}
\end{minipage}
\end{figure}

Fig.~\ref{fig:dp_loss} shows the held-out loss during DP training. With $\sigma=0$, roughly matches held-out loss after about \num{1200} steps. When training with $\sigma=2^{-8}$, equivalent held-out loss is achieved after \num{2700} steps. This corresponds to an analytical $\mu$-GDP guarantee of roughly \num{13000}.\footnote{Because we are using Poisson sampling this bound relies on the pessimistic assumption that the example participates in every iteration; we get a more realistic $\mu$-GDP bound of $\approx 572$ by assuming each example participates five times.} Fig.~\ref{fig:dp_5epochs} demonstrates that under this model achieving comparable utility with a weak DP guarantee, our synthetic canaries are still able to detect more memorization than the baselines.


On the other hand, one might accept a hit to utility in order to achieve higher privacy by stopping DP training earlier. We look at the MI results of $\sigma=0$ and $\sigma=2^{-8}$ after just 1350 steps of training (Fig.~\ref{fig:dp_2p5epochs}), at which point the model trained with $\sigma=0$ achieves comparable utility to our baseline. Here the analytical $\mu$-GDP parameter is \num{9300}, with the more realistic value corresponding to some examples participating 3 times being about \num{443}. Comparing the MI results of the DP models trained with $\sigma=0$ and $\sigma=2^{-8}$, we see that even a minimal amount of DP noise significantly reduces memorization. 


Notably, for all of these DP settings, only our synthetic canaries indicate a concerning amount of leakage: empirical $\mu > 1.35$. This threshold corresponds to 75\% TPR-at-25\%-FPR, or $\varepsilon = \ln(3)$. Auditing using any existing canary type would have failed to demonstrate that this noise level is not safe to use.


\subsection{Model Quality Assessments}
\label{apx:model_quality}

\subsubsection{Loss} 
\begin{figure}[htb]
\centering
\begin{minipage}{.48\textwidth}
  \centering
  \includegraphics[width=1.1\linewidth]{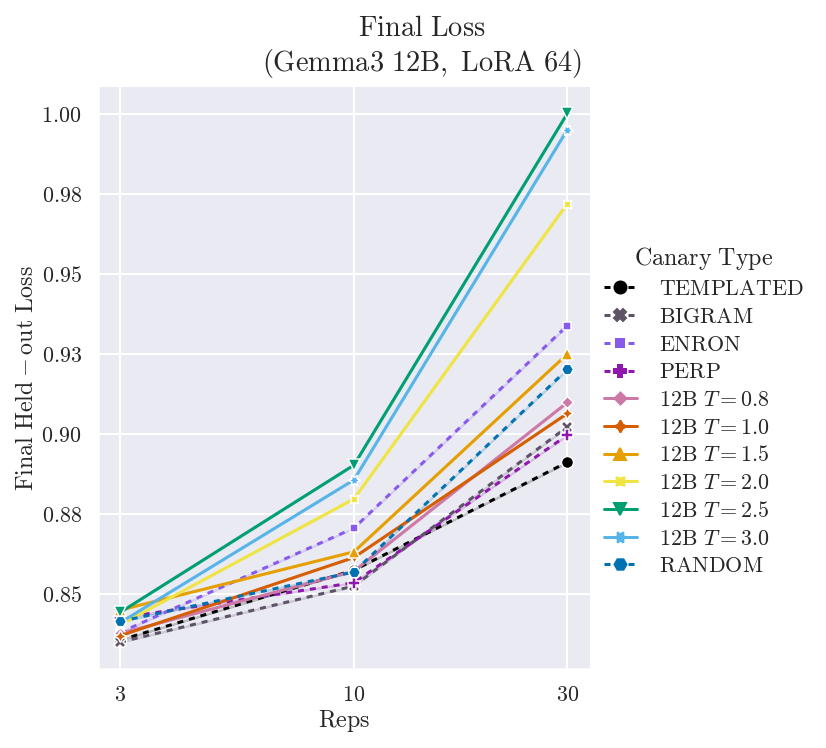}
    \caption{At high numbers of repetitions, the inclusion of especially high-temperature synthetic canaries can slightly increase the final fine-tuned model's held-out loss slightly more than other canary types.}
    \label{fig:loss}
\end{minipage}\hfill
\begin{minipage}{.48\textwidth}
  \centering
  \includegraphics[width=.9\linewidth]{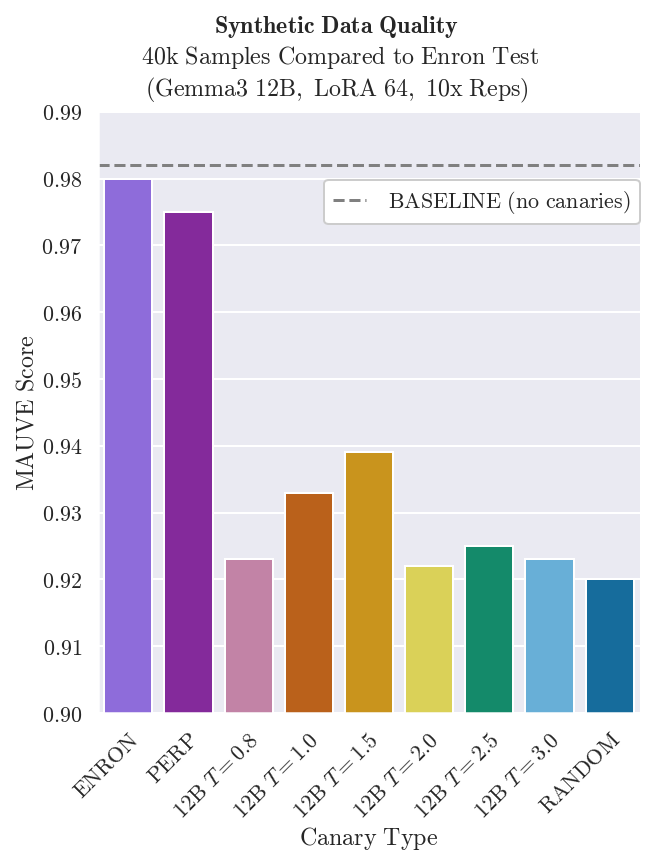}
    \caption{MAUVE scores of the synthetic data generated from models fine-tuned with 10x reps of different canaries, measured against held-out Enron test examples. Synthetic canaries do not degrade the resulting synthetic data quality any more than other canary types that do not rely upon access to the private data.}
    \label{fig:mauve}
\end{minipage}
\end{figure}

We find that the final held-out loss after a fine-tuning on Enron with canaries is somewhat impacted by the inclusion of canaries. Fig.~\ref{fig:loss} shows how high numbers of reps of high temperature $T\ge2$ canaries can degrade the final model loss.

\subsubsection{MAUVE}
\label{apx:mauve}
An important application of fine-tuning on sensitive data is for synthetic data generation~\citep{ponomareva2025dpfydata}. Private synthetic data can be used in place of real data for many purposes. However the presence of out-of-distribution canaries during training can negatively impact synthetic data quality by causing the generation process to produce examples that resemble the canaries instead of the real data. We compared the impact of canaries during training on synthetic data quality by evaluating the MAUVE score~\citep{pillutla2021mauve} between 40k synthetic examples generated from the fine-tuned model and the Enron test data. Test examples that are used for ENRON canaries and for generating PERP canaries are excluded from this set.

The MAUVE metric~\citep{pillutla2021mauve} is highly effective for the purpose of rigorously quantifying the utility of the generated data. It evaluates the gap between the true data distribution ($P$) and the synthetic data distribution ($Q$) by computing the area under the Information Divergence Frontier. Rather than relying on simple token overlap, MAUVE captures the distributional trade-off between precision (how closely the synthetic text resembles the true text) and recall (how well the synthetic text covers the diversity of the true data) using quantized embeddings from a pre-trained language model. 

We compute MAUVE using the 308M parameter EmbeddingGemma model\footnote{\href{https://ai.google.dev/gemma/docs/embeddinggemma}{\texttt{https://ai.google.dev/gemma/docs/embedding\\gemma}}} to generate embeddings for the text samples. The embeddings are reduced in dimensionality using PCA to retain 90\% of the variance. The MAUVE score is then calculated following the recipe from \citet{pillutla2021mauve}, which discretizes the embedding space using k-means clustering. The number of clusters is set to 10\% of the total number of samples, clamped between 2 and 500, as suggested in the original MAUVE implementation.

Fig.~\ref{fig:mauve} contains results on the synthetic data generation quality of our Gemma3 12B models fine-tuned on Enron via LoRA 64 across canary types, repeated 10 times. As a baseline we consider the MAUVE score of synthetic data generated from the Gemma3 12B model fine-tuned on Enron without canaries. We find, unsurprisingly, that held-out examples from Enron test have the least impact on the resulting generated model quality. PERP canaries~\citep{meeus2025canarysecho} also perform especially well. This is due to the fact that PERP canaries are generated by using prefixes from the fine-tuning distribution. Constructing both ENRON and PERP canaries require access to the private fine-tuning data, a constraint that may be prohibitive in practice. While our synthetic canaries hurt MAUVE score some, when compared to RANDOM canaries that also do not rely on private data, our synthetic canaries offer an advantage on resulting model synthetic data quality. The effect would presumably decrease if canaries were inserted at fewer than 10 repetitions in a real auditing scenario.



\subsection{Additional Data Audit Results}

\subsubsection{Comparing Data Audit Approaches}
\label{apx:bigram}

Table~\ref{tab:bigram_comparison} contains the complete MI metrics comparing our LLM-based data audit attack to the bigram data audit attack of \citet{meeus2025canarysecho}. We find that across canary types and repetitions, and regardless of the metric, our LLM-based approach far surpasses the signal produced by the bigram model attack. Note that we train a single model (bigram and LLM, respectively) to conduct our attacks, whereas \citet{meeus2025canarysecho} report their results using an RMIA-style attack that requires generating multiple synthetic datasets from models with different sets of canaries, and training bigram models on each of them~\citep{zarifzadeh2024lowcost}. Our approach presents a more realistic attack where the adversary has access to the actual released set of synthetic data. Running RMIA on bigram and LLM-based data audits should boost the MI performance of both approaches. Even in this more challenging setting of only using a single finetuning run to conduct the data audit, our LLM-based data audit achieves significant MI performance. It is also interesting to note that if the weaker Bigram attack is used, the PERP canaries do outperform the temperature-based canaries, perhaps because the canaries were designed to be strong with that attack.

\begin{table*}[t]
\centering
\caption{A direct comparison of our LLM-based data audit with the bigram data audit of~\citet{meeus2025canarysecho} across MI metrics for varying canary types and repetitions. Neither attack employs RMIA, which would increase the effectiveness of each attack, at the expense of computation and realism. As ENRON and RANDOM canaries performed poorly at 10 repetitions, we elected to not expend the compute to evaluate them at 3 repetitions.}
\label{tab:bigram_comparison}
\begin{tabular}{ll|cc|cc|cc}
\toprule
Reps & Canary & \multicolumn{2}{c|}{$\mu$} & \multicolumn{2}{c|}{TPR@0.01FPR} & \multicolumn{2}{c}{TPR@0.1FPR} \\
& \ Type & Bigram & LLM & Bigram & LLM & Bigram & LLM \\
\midrule
10 & PERP & \textbf{0.266} & 2.037 & \textbf{0.0183} & 0.1723 & \textbf{0.137} & 0.394 \\
10 & $T=0.8$ & 0.133 & 1.869 & 0.0116 & 0.1717 & 0.105 & 0.402 \\
10 & $T=1.0$ & 0.231 & 2.229 & 0.0144 & 0.1797 & 0.116 & 0.443 \\
10 & $T=1.5$ & 0.109 & 2.210 & 0.0126 & 0.2076 & 0.108 & 0.486 \\
10 & $T=2.0$ & 0.161 & 2.501 & 0.0149 & \textbf{0.3270} & 0.112 & \textbf{0.645} \\
10 & $T=2.5$ & 0.146 & 2.410 & 0.0127 & 0.2839 & 0.108 & 0.580 \\
10 & $T=3.0$ & 0.183 & \textbf{2.519} & 0.0146 & 0.2434 & 0.107 & 0.529 \\
10 & ENRON & 0.225 & 1.448 & 0.0139 & 0.1232 & 0.106 & 0.315 \\
10 & RANDOM & 0.200 & 0.609 & 0.0134 & 0.0242 & 0.109 & 0.171 \\
\midrule
3 & PERP & \textbf{0.213} & 0.390 & \textbf{0.0155} & 0.0206 & \textbf{0.130} & 0.160 \\
3 & $T=0.8$ & 0.104 & \textbf{0.699} & 0.0109 & \textbf{0.0280} & 0.104 & 0.166 \\
3 & $T=1.0$ & 0.206 & 0.482 & 0.0118 & 0.0278 & 0.112 & 0.174 \\
3 & $T=1.5$ & 0.107 & 0.568 & 0.0130 & 0.0268 & 0.111 & \textbf{0.174} \\
3 & $T=2.0$ & 0.162 & 0.364 & 0.0150 & 0.0239 & 0.110 & 0.143 \\
3 & $T=2.5$ & 0.142 & 0.307 & 0.0124 & 0.0179 & 0.105 & 0.131 \\
3 & $T=3.0$ & 0.194 & 0.260 & 0.0141 & 0.0183 & 0.107 & 0.131 \\
\bottomrule
\end{tabular}
\end{table*}


\end{document}